\documentclass[10pt,twocolumn,letterpaper]{article}

\usepackage{cvpr}
\usepackage{times}
\usepackage{graphicx}
\usepackage{amsmath}
\usepackage{amssymb}
\usepackage{tikz}
\usetikzlibrary{babel}

\usepackage[pagebackref=true,breaklinks=true,colorlinks,bookmarks=false]{hyperref}

\cvprfinalcopy 

\ifcvprfinal\fi

\begin{document}

\title{FloorPP-Net: Reconstructing Floor Plans using Point Pillars for Scan-to-BIM}

\author{Yijie Wu \qquad Fan Xue\\
The University of Hong Kong\\
Pokfulam, Hong Kong, China\\
{\tt\small \{wuyijie, xuef\}@hku.hk}
}

\maketitle

\begin{abstract}
This paper presents a deep learning-based point cloud processing method named FloorPP-Net for the task of Scan-to-BIM (building information model)~\cite{bosche2015value}. FloorPP-Net first converts the input point cloud of a building story into point pillars (PP) \cite{lang2019pointpillars}, then predicts the corners and edges to output the floor plan. Altogether, FloorPP-Net establishes an end-to-end supervised learning framework for the Scan-to-Floor-Plan (Scan2FP) task. In the 1st International Scan-to-BIM Challenge (\url{https://cv4aec.github.io/}) held in conjunction with CVPR 2021, FloorPP-Net was ranked the second runner-up in the floor plan reconstruction track. Future work includes general edge proposals, 2D plan regularization, and 3D BIM reconstruction.
\end{abstract}

\section{Introduction}

Floor plan is an essential and popular representation of building interiors in the Architecture, Engineering, and Construction (AEC) industry. With advanced technologies, such as building information model (BIM) and high-definition point cloud 3D scanning, Scan-to-BIM is the task of reconstructing accurate and detailed BIMs from point cloud scans~\cite{bosche2015value}. The reconstruction of floor plan (Scan2FP) is thus highly desired by Scan-to-BIM and AEC practitioners. However, Scan-to-BIM (and Scan2FP) encounters computational challenges, including large-scale inputs, data diversity (e.g., different indoor scenarios), accurate geometry, and detailed semantics (e.g., labels of building components)~\cite{xue2021semantic}. A general automatic solution must handle the challenges in order to achieve satisfactory generalizability, effectiveness, and efficiency. 

In recent years, deep neural networks (DNNs) showed promising potentials in many tasks of parsing geometry and semantics. The DNNs have also been applied to solving several Scan2FP problems. For example, FloorNet \cite{liu2018floornet} takes the point clouds and images as inputs and uses three separate branches for extracting the geometry and semantics from 3D, point density of top-down view, and images to predict the pixel-wise floor plans. Floor-SP \cite{chen2019floor} uses Mask R-CNN \cite{he2017mask} to segment rooms from the density/normal maps of the top-down view and then optimizes the boundary loops of rooms. These works proposed novel frameworks to learn Scan2FP from data and gained state-of-the-art outcomes on the residential datasets. 
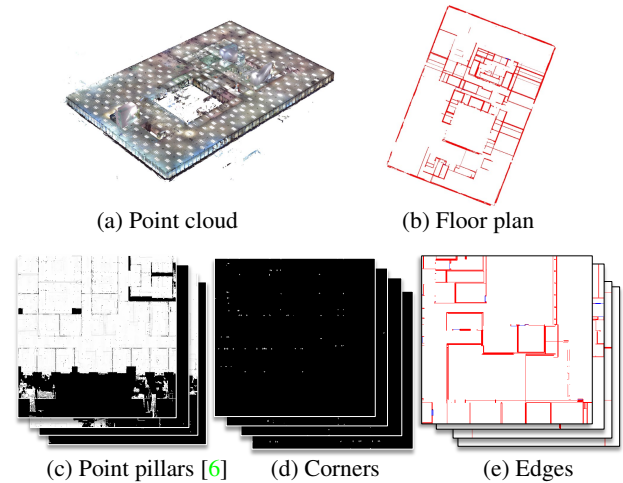
\begin{figure}[t]
\centering
\begin{tikzpicture}
		\draw (0, 0) node {\includegraphics[width=1\linewidth]{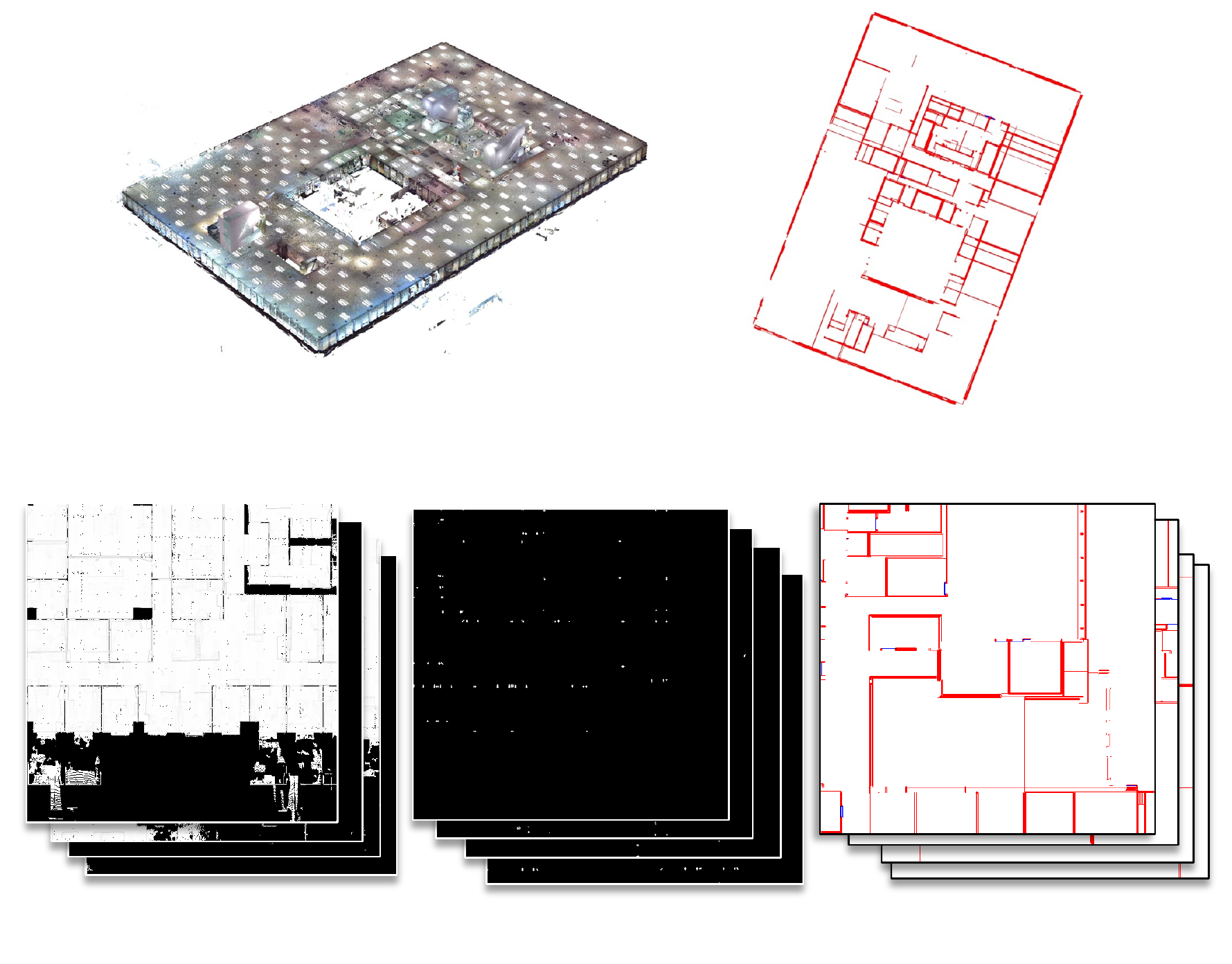}};
		\draw (-2, 0.3) node {\small (a) Point cloud};
        \draw (2, 0.3) node {\small (b) Floor plan};
	    \draw (-2.4, -3) node {\small(c) Point pillars \cite{lang2019pointpillars}};
	    \draw (0.1, -3) node {\small(d) Corners};
		\draw (2.8, -3) node {\small(e) Edges};
	\end{tikzpicture}
   \caption{Our method takes the (a) point cloud of a building story as input, 
   then converts the point cloud into (c) point pillars of the corresponding 2D horizontal grid. 
   Next, FloorPP-Net learns to predict the (d) corners and (e) edges 
   to output the (b) final floor plans.}
\label{fig:long}
\label{fig:onecol}
\end{figure}

\begin{figure*}[t]
\centering
\includegraphics[width=0.9\linewidth]{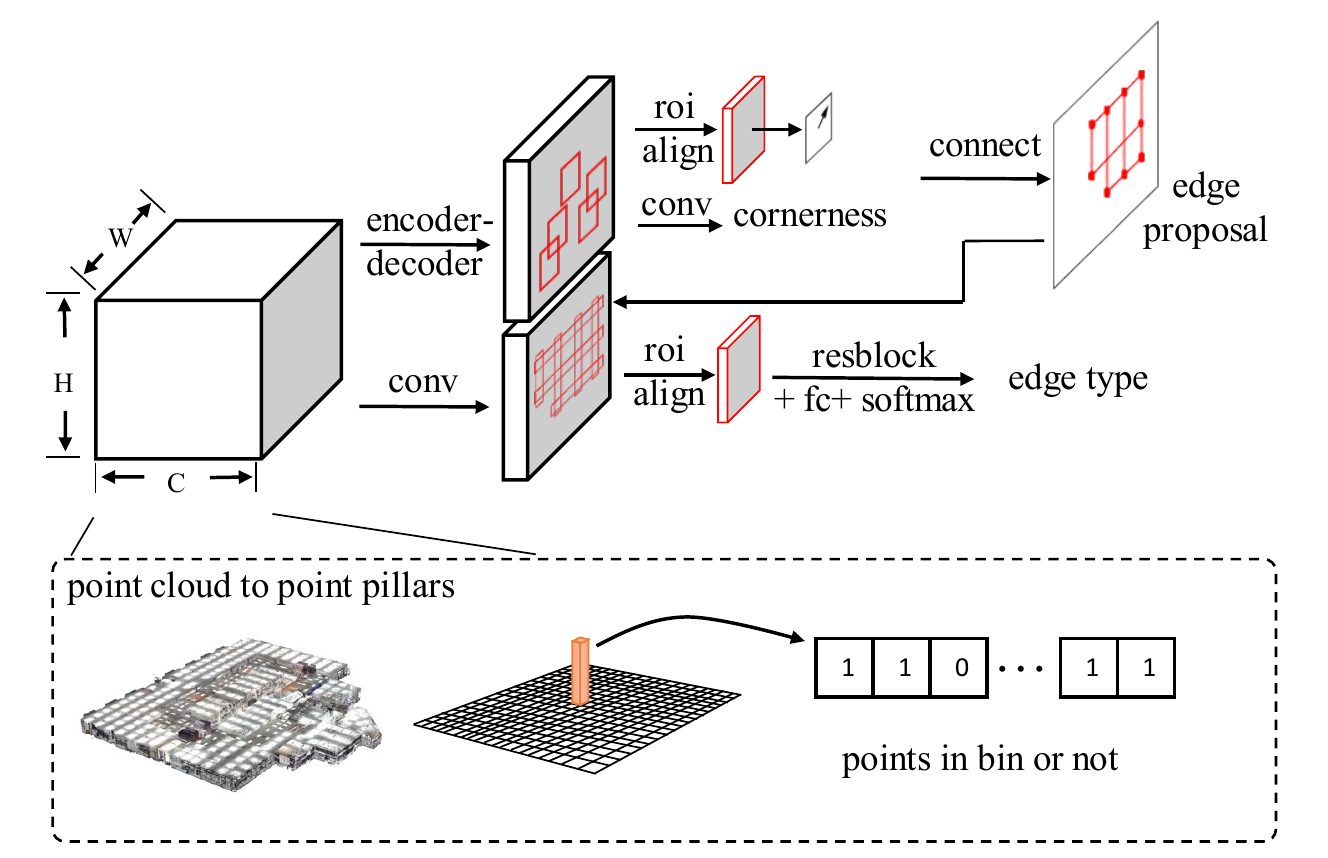}
   \caption{Overview of FloorPP-Net.}
\label{fig: overview}
\end{figure*}

In this paper, we propose a method named FloorPP-Net. FloorPP-Net converts the point cloud into point pillars (PP) \cite{lang2019pointpillars} and learns to predict the corners and edges from the point pillars to form the floor plan. A point pillar in this paper is a compact geometric feature from a top-down view. In general, the point pillar is more distinguishable and robust than the conventional point density/normal maps. Moreover, by simultaneously learning the corners and edges, FloorPP-Net is an end-to-end pipeline for Scan2FP. Our multi-task learning framework for corners and edges is similar to the LCNN \cite{zhou2019end}. However, FloorPP-Net exploits the innate architectural regularities~\cite{wu2021regard}, thus is able to simplify the edge sampling and proposals. 

FloorPP-Net was ranked the second runner-up in the floor plan reconstruction track in the 1st International Scan-to-BIM Challenge held in conjunction with CVPR 2021. The results showed the general framework of FloorPP-Net is reasonable; yet, some components of FloorPP-Net were still in their naive versions and some components were not implemented by the competition. For example, the architecture and hyper-parameters of FloorPP-Net were not yet fully tuned.

\section{FloorPP-Net}
\subsection{Overview}
Figure~\ref{fig: overview} shows the pipeline of the FloorPP-Net. First, the input point clouds of building stories are preprocessed into point pillars, as shown in Figure~\ref{fig: overview}b and Section~\ref{sec: prep}. Next, FloorPP-Net takes the point pillars as inputs and detects the corners through an encoder-decoder with an ROI detection head (Section~\ref{sec: corner}). Once corners are detected, FloorPP-Net connects corners, generates edge proposals, and classifies edges (Section~\ref{sec: edge}). As shown in Figure~\ref{fig: overview}, FloorPP-Net has separate heads for corner detection and edge verification. Hence, the loss function of FloorPP-Net is the weighted sum of the loss $L_C$ for corner detection and $L_E$ for edge verification, which is $L = L_C + \lambda_EL_E$, where $\lambda_E$ is the weight parameter.

\subsection{Point cloud to point pillars}\label{sec: prep}
To convert the point cloud into point pillars, FloorPP-Net employs a 2D grid for the horizontal bounding box. Feature vectors are computed within the 2D grid cells to represent the vertical point distributions along the corresponding vertical prism, or pillars. Specifically, we compute the feature vector in two steps:
\begin{enumerate}
\item binning a pillar at equal intervals and 
\item labeling the bins as TRUE if points inside and FALSE otherwise.
\end{enumerate}

The feature computation is simple and intuitive. Yet the feature is sensitive to the vertical outliers. Because the lengths of feature vectors, i.e., the numbers of vertical bins, are fixed for all point cloud samples, points belong to the target story will be squeezed into few bins when far outliers exist. To alleviate this negative effect, we estimated the elevations of each story's floor and ceiling and removed the points outside of the vertical range of a story. Furthermore, the point clouds are aligned to the x- and y-axes to reduce the rounding error~\cite{wu2021regard}.

\begin{table*}
  \centering
  \caption{Official metrics report.}
    \begin{tabular}{l|ccccccccc}
    \hline
         Method & \multicolumn{1}{l}{warping\_error} & \multicolumn{1}{l}{precision\_0} & \multicolumn{1}{l}{precision\_1} & \multicolumn{1}{l}{precision\_2} & \multicolumn{1}{l}{recall\_0} & \multicolumn{1}{l}{recall\_1} & \multicolumn{1}{l}{recall\_2} & \multicolumn{1}{l}{IoU} & \multicolumn{1}{l}{betti\_error} \\
   \hline\hline
    FloorPP-Net & 0.268 & 0.011 & 0.042 & 0.065 & 0.071 & 0.256 & 0.386 & 0.120 & 1.204 \\
    \hline
    \end{tabular}%
  \label{tab: metric}%
\end{table*}

\subsection{Corner detection}\label{sec: corner}
FloorPP-Net utilizes an encoder-decoder structure to produce a feature map with high resolution and rich context information for detecting corners. The corner scores of grid cells are predicted on the feature map. However, in pilot experiments, we observed that the `per-pixel' corner prediction produced duplicated and blurred results as well as caused noisy edge proposals. To focus on appropriate corner proposals and results, FloorPP-Net refines the corner locations using the RoIAlign \cite{he2017mask} on the grid cells with higher corner scores. Therefore, the loss function of the corner detection includes two parts: 
$$L_C = L_{cls} + \lambda_{loc} L_{loc},$$
where $L_{cls}$ is the binary cross-entropy loss over two classes: corner or not; $L_{loc}$ is a smooth $L_1$ function \cite{girshick2015fast} that measures the errors between the refinement and the ground truth location; $\lambda_{loc}$ is a weight parameter to balance the two parts of the $L_C$.

To ease the learning of the corner detection, $L_C$ only takes the filtered grid cells into account. We use a similar filtering strategy as Faster R-CNN \cite{ren2015faster}:
\begin{enumerate}
\item generating boxes of each grid cell and ground-truth corners with a given side length;
\item computing the box Intersection-over-Union (IoU) between the grid cells and ground-truth corners; and
\item selecting those cells with a maximum IoU larger than 0.7 as positive samples and lower than 0.3 as negative samples.
\end{enumerate}
The cells with maximum IoU from 0.3 to 0.7 are ignored in the computation of $L_C$. Meanwhile, since the number of the negative cells can still be overwhelming to the positive, we randomly sample from the negative cells to keep a balanced ratio of positive to negative. Besides, only the location refinement of the positive samples contributes to the $L_{loc}$.

\subsection{Edge verification}\label{sec: edge}
After the corner detection, FloorPP-Net generates edge proposals by connecting the detected corners with refined locations. Although an intuitive way to propose edges is filtering the full connections between all the detected corners. However, the full connection leads to a large number of proposals, and an extremely imbalanced ratio of positive to negative samples in particular. Thanks to the axis alignment in the preprocessing, the true positive edges' orientations are highly regular on common Manhattan-like floor plans. Therefore, simple heuristic rules can be applied to connect corners in a very efficient way. Based on the Manhattan layout assumption for the floor plans, FloorPP-Net only connects horizontal and vertical edges and ignores the unlikely or inapplicable connections. The main side effect from the Manhattan assumption was the missing of incline edges. 

Once the edge proposals are generated, the RoIAlign \cite{he2017mask} predicts the edge score for each proposal. We use the binary cross-entropy loss as $L_E$. Besides, FloorPP-Net employs a filter, similar to that of the corner detection, of the edge proposals, to reduce the complexity of learning.

\section{Implementation and Results}
FloorPP-Net was initially implemented for the 1st International Scan-to-BIM Challenge (\url{https://cv4aec.github.io/}). We adopted ResNet-18 as the backbone of the encoder-decoder structure of the corner detection. Three residual blocks were stacked to extract the feature maps of the edge proposals. We trained the FloorPP-Net model from scratch for 55 epochs with a learning rate of 0.0001 which was decreased by 10 at epoch 40. The FloorPP-Net model was trained on a GPU with 8 GB memory. The batch size for training was set to 1. All point clouds were down-sampled to voxels at the resolution of 0.05m. Every point cloud was cropped into horizontal grids with a side length of 512; while the side length of the corner box was set to 9.

Table~\ref{tab: metric} shows the error metrics computed by the Challenge organizers. Overall, FloorPP-Net won the 2nd runner-up in the floor plan reconstruction track. The 0.268 warping error~\cite{jain2010boundary} indicated FloorPP-Net returned floor plans with low homotopic deformations. Yet, the precision-recall pairs were not satisfactory. E.g., all the three values of precision were below 0.1, and the IoU was only 0.120. The indications and details of the metrics can be found on the Challenge web pages at \url{https://cv4aec.github.io/} and \url{https://github.com/seravee08/WarpingError_Floorplan}.

\section{Discussion and future work}
The method FloorPP-Net proposed in this paper focuses on corners, a kind of typical joints in building interiors, for the task of Scan2FP and Scan-to-BIM. The point pillars (PPs) and edges between PPs are the key features to handle in the FloorPP-Net model. Results showed that the PPs and PP-based edges were reasonable and efficient to process large-scale point cloud scans, e.g., 200 million points of a building floor with more than 100 rooms.

There are three major directions to complete and improve FloorPP-Net. The first is to include the non-Manhattan edges between PPs. Another one is the regularization of walls and spaces (e.g.,~\cite{xue2020lidar}) for floor plans. Subsequently, 3D as-built BIMs can be reconstructed for Scan-to-BIM based on confident 2D floor plans.

{\small
\bibliographystyle{ieee}
\bibliography{egbib}
}

\end{document}